\documentclass[sigconf, nonacm]{acmart}

\newcommand{\design}{PolyQ}

\usepackage{hyperref}
\usepackage{amsmath,amsfonts}
\usepackage{algorithm,algpseudocode}
\algnewcommand\algorithmicforeach{\textbf{for each}}
\algdef{S}[FOR]{ForEach}[1]{\algorithmicforeach\ #1\ \algorithmicdo}
\usepackage{listings}
\usepackage{array}
\usepackage{graphicx}
\usepackage{textcomp}
\usepackage{xcolor}
\usepackage{orcidlink}
\usepackage{svg}
\usepackage{enumitem}
    
\usepackage{multirow}
\usepackage{makecell}
\usepackage{booktabs}

\usepackage[font=footnotesize]{caption}

\begin{document}

\title{\design: Codesigning End-to-End Quantization Framework for Scalable Edge CPU LLM Inference}

\author{Hyunwoo Oh, Suyeon Jang, Hanning Chen, KyungIn Nam, Sanggeon Yun,\\Ryozo Masukawa, and Mohsen Imani}
\affiliation{\institution{University of California, Irvine}
\country{}}
\email{{hyunwooo, m.imani}@uci.edu}

\begin{abstract}
CPUs are the most universal target for on-device LLM inference, but existing low-bit quantization methods offer either coarse operating points or fine-grained mixed precision that is difficult to execute efficiently on CPUs.
We present \design{}, a CPU-oriented compiler/quantization co-design for activation-aware channel-wise bit allocation under a user-specified average-bit budget.
\design{} assigns per-channel bit-widths from $\{2,3,4,8,16\}$, then uses a compile-time model compiler to permute and cluster channels into bit-homogeneous blocks, generate SIMD- and LUT-compatible kernels, and merge compatible permutations across operators to keep layout regularization off the runtime path.
This turns fine-grained budget fitting into a practical fractional-bit deployment method for CPU-only inference.
Across Falcon-H1-3B, Llama2-13B, and Qwen3-32B on WikiText-2, \design{} provides stable quality scaling from 3--6\,b and improves perplexity by 2.4--32.1\% over prior methods at a 3\,b target.
End-to-end measurements on three representative CPUs---workstation, laptop, and mobile---show that compiler layout regularization reduces activation reorder traffic by up to 70.8\%, prefill latency and decode throughput scale 
nearly proportionally with the configured bit budget, and energy/token overhead stays below 2\% relative to an optimized LUT-based back-end.
These results show that fractional-bit CPU deployment is practical, predictable, and energy-efficient across diverse edge targets.
\end{abstract}

\maketitle

\section{Introduction}
Large language models (LLMs) are increasingly expected to run on-device---on laptops, edge servers, and embedded controllers~\cite{edge_llm_survey, ondevice0, ondevice1, ondevice2}---favoring privacy, low latency, and cost.
CPUs are universally present across these targets, operate under tight power and cost budgets, and support simple software stacks.
Although GPUs and accelerators offer excellent peak throughput \cite{oltron}, many edge workloads run at batch$\,{=}\,1$ \cite{gcvturbo, hummingbird}, where parallel compute is under-utilized and the deployment advantage over CPUs diminishes~\cite{llmflash}, making CPUs the dependable default.

Edge deployments face memory constraints more nuanced than a single model-size limit:
usable memory is fragmented by co-running software and shared clients, while weights compete with KV cache, activations, and the software stack.
Batch$\,{=}\,1$ inference is dominated by weight storage and DRAM traffic~\cite{ecco, squeezellm}, making coarse operating points such as W3 or W4 inefficient---they either leave slack unused or overshoot the budget.
This makes flexible budget planning---not just low precision itself---a first-class systems requirement.

Prior PTQ methods reduce model size and memory traffic without retraining~\cite{evalquant, smoothquant, gptq}, and inter-layer mixed-precision methods extend this with coarse budget control~\cite{haq, amq}. Recent activation-aware \cite{awq} and intra-layer methods exploit within-layer saliency for finer adaptation~\cite{slimllm}, reordering and compiler methods improve execution efficiency~\cite{rptq, atom}, and LUT-based kernels demonstrate efficient sub-byte execution on CPUs~\cite{tmac, bitnet_cpp, tsar}.
Despite this progress, practical CPU-only edge deployment remains unresolved. Three issues are central to this gap:

\textbf{1. Deployable fine-grained controllable PTQ method.}
Inter-layer methods~\cite{amq} limit budget granularity and palette to 2/3/4-bit, preventing high-precision protection of activation-salient outlier channels and restricting deployable budget range.
Intra-layer methods~\cite{slimllm} offer finer allocation but adopt $b\pm$1-bit entries that only support coarse integer-wise budget control and force INT8 dequantization on CPU-incompatible 5--7-bit entries that forfeit LUT efficiency due to $2^b$ table scaling, while still lacking the 8/16-bit range needed to match the diverse saliency distribution.
What is missing is a method with a CPU-aligned palette spanning 2--16-bit, enabling both saliency-matched allocation and fractional-bit budget control.

\textbf{2. Efficient layer-local kernels for heterogeneous low-bit blocks.}
Prior CPU kernels target uniform low-bit configurations~\cite{tmac, bitnet_cpp, tsar}, and mixed-precision approaches rely on requantization around GEMM~\cite{atom}, introducing conversion and data-movement overhead. Translating fine-grained bit assignments into efficient heterogeneous low-bit execution on CPUs remains difficult.

\textbf{3. Efficient graph-level layout regularization.}
Local clustering and reordering~\cite{rptq, atom} reduce within-layer overhead, but permutations between adjacent operators reintroduce runtime data movement. A compile-time layout-regularization path is needed to convert irregular channel-wise bit allocations into bit-homogeneous execution blocks without runtime permutation cost.

These challenges limit the practicality of CPU-only inference and motivate a design that jointly addresses flexible budget planning, heterogeneous low-bit execution, and low-overhead permutation.

This paper closes the gap between coarse but deployable inter-layer control and fine-grained but execution-unfriendly intra-layer mixed precision.
We present \textbf{\design{}}, a CPU-oriented framework for deployable activation-aware channel-wise mixed-precision inference.
\design{} assigns per-channel bit-widths from a CPU-aligned \{2,3,4,8,16\}-bit palette under a user-specified fractional average-bit budget based on activation-aware saliency, then regularizes the resulting layout at compile time through permutation, clustering, and layer-specific kernel generation.
By propagating compatible permutations across operators, \design{} keeps most layout transformations off the runtime path.
This decouples fine-grained budget adaptation from regular kernel execution, enabling deployable intra-layer mixed precision whose runtime scaling tracks the bit budget.

The contributions are as follows:
\begin{itemize}[leftmargin=*, nosep]
\item \textbf{A deployable channel-wise mixed-precision formulation.}
We propose an intra-layer quantization method over a CPU-aligned \{2,3,4,8,16\}-bit palette, guided by activation-aware saliency, enabling 
saliency-matched bit allocation and fine-grained fractional-bit budget adaptation across diverse deployment targets.

\item \textbf{A compile-time co-design for regular CPU execution.}
We present a compiler that clusters irregular channel-wise bit assignments into bit-homogeneous blocks, generates matching low-bit CPU kernels, and propagates compatible permutations across operators to reduce runtime layout overhead.

\item \textbf{End-to-end validation of budget scalability and efficiency.}
We show that \design{} improves perplexity by 2.4--32.1\% over prior methods at a 3b target, realizes fractional budgets within 0.045b of target after SIMD-optimized quanta matching, reduces activation reorder traffic by up to 70.8\% over prior layout policies, and keeps end-to-end latency within 5.8\% of an optimized LUT backend with less than 2\% energy/token overhead.
\end{itemize}
\section{Background, Related Works, and Motivation}
\label{sec:bg}

On-device LLMs on CPUs are typically constrained by weight storage and DRAM bandwidth.
Conventional PTQ reduces this footprint, but the control surface is coarse: one picks a single bit-width per layer (e.g., W3 or W4) to fit a memory and latency envelope.
This works well when deployment budgets align with the low-bit ladder, but becomes inefficient for fractional targets such as 3.8\,bits/weight.
This problem is particularly visible when the usable memory region is limited, fragmented or tightly bounded by co-running software: coarse W3/W4 operating points often either leave too much slack or overshoot the envelope.

\begin{figure}[t]
\centering
\includegraphics[width=0.93\linewidth]{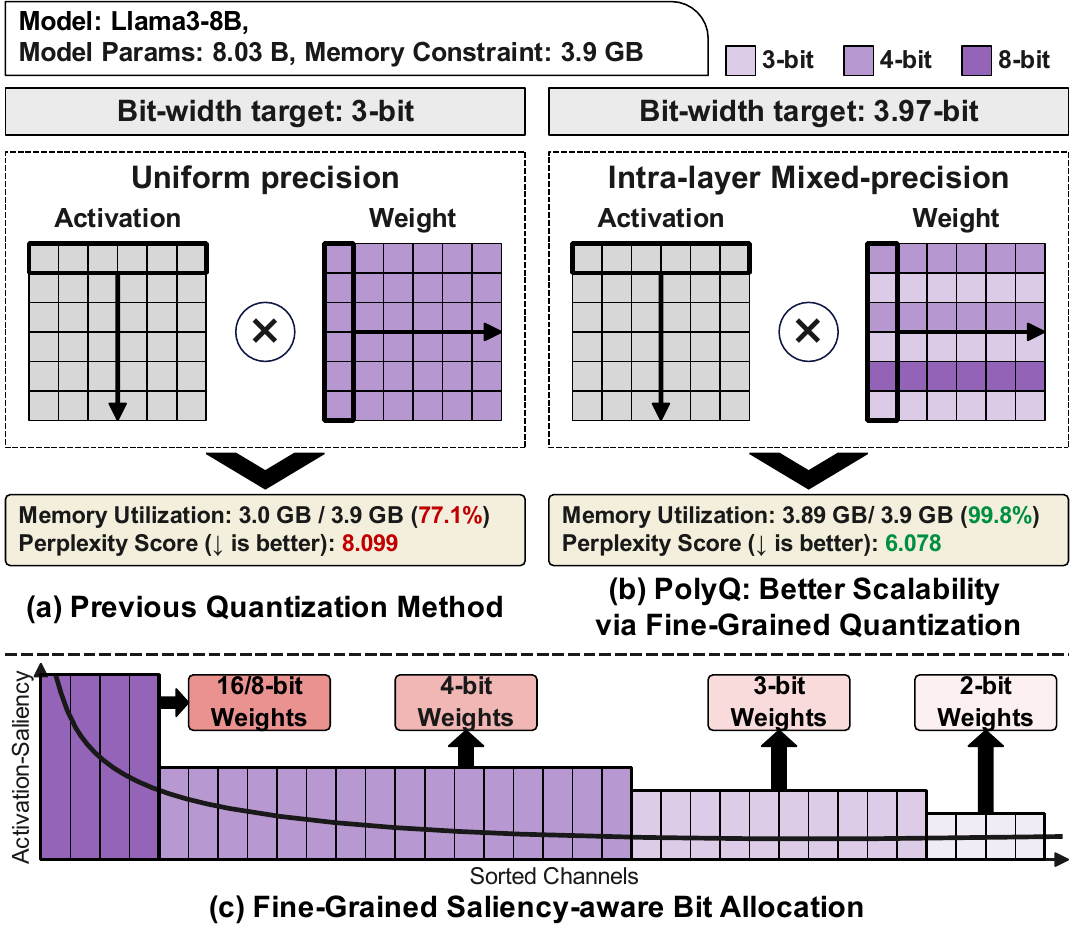}
\caption{
Budget-fit motivation for fine-grained quantization on CPUs, benchmarked on WikiText-2 with Llama3-8B under a 3.9\,GB memory budget.
Panels (a) and (b) visualize channels in activation--weight space and the bit-width assigned to them.
(a) A conventional uniform-precision scheme targeting 3\,bits/weight leaves substantial headroom (3.0\,GB used, 77.1\% of the budget) and suffers higher perplexity (8.099).
(b) \design{} assigns bits per channel and achieves 3.97\,bits/weight by mixing 2/3/4/8/16-bit channels, using nearly the full budget (3.89\,GB, 99.8\%) and significantly reducing perplexity (6.078).
(c) A schematic view of the same allocation principle: after channels are sorted by activation saliency, higher precision is reserved for a small salient subset while most channels remain at lower precision.
}
\label{fig:motivation}
\end{figure}

\autoref{fig:motivation} illustrates the problem and the opportunity.
Assuming we assign a 3.9\,GB memory budget for Llama3-8B on WikiText-2, \autoref{fig:motivation}(a) shows that a uniform layer-wise 3-bit point underfills the budget and yields worse (higher) perplexity score, whereas \autoref{fig:motivation}(b) shows that \design{} allocates bits \emph{per channel}, reaches near-budget fit, and substantially improves perplexity.
\autoref{fig:motivation}(c) then gives the intuition behind this gain: higher precision is concentrated on a small set of activation-salient channels, while most channels remain low-bit.
The lesson is therefore not merely \emph{how many} bits we use on average, but \emph{where} those bits are spent and how closely the resulting footprint matches the target envelope.

\noindent\textbf{Activation-aware and uniform methods.}
AWQ~\cite{awq} shows that only a small fraction of salient channels dominate the layer output and deserve protection, applying activation-aware scaling to improve robustness at a fixed nominal bit-width.
SmoothQuant~\cite{smoothquant} and related work~\cite{pushquant} similarly highlight the importance of activation-sensitive error control.
However, these methods retain a uniform per-layer bit-width: the activation signal identifies \emph{where} precision matters, but does not translate into a flexible budget control method.
GPTQ~\cite{gptq} achieves strong low-bit compression through second-order weight update but likewise targets fixed integer operating points with no fractional-budget control.

\noindent\textbf{Inter-layer mixed-precision methods.}
HAQ~\cite{haq} and AMQ~\cite{amq} expose budget control by assigning different bit-widths across layers, aligning naturally with efficient execution---one kernel path per layer---but the granularity is structurally limited. At layer level, protecting a salient subset requires elevating the entire layer's 
precision, which is wasteful, and a 2/3/4-bit palette ceiling means highly salient outlier channels cannot be adequately protected without overprovisioning the entire layer.
Approximating fractional targets requires exploring Pareto-efficient layer-wise assignments, which is computationally expensive and remains imprecise due to the coarse layer-level granularity.

\noindent\textbf{Intra-layer mixed-precision methods.}
Slim-LLM~\cite{slimllm} pushes to channel group-level granularity, exploiting within-layer saliency for finer accuracy--memory trade-offs.
However, its b$\pm$1 palette includes 5--7-bit entries that are structurally incompatible with LUT-based CPU kernels: LUT sizes scale as $2^b$, making 5/6/7-bit entries require 32/64/128-entry tables that break SIMD efficiency, forcing dequantization to INT8 and forfeiting the throughput and energy benefits of low-bit execution.
This also restricts Slim-LLM to coarse integer-bit targets with no fractional-budget control.

\noindent\textbf{Execution-side: reordering and compiler methods.}
RPTQ \cite{rptq} shows that fixed channel reordering within a transformer block can improve quantization robustness and reduce runtime overhead.
Atom \cite{atom} similarly uses mixed-precision grouping and reordering to improve GEMM efficiency at operator boundaries.
Both are important steps, but they address only local reordering within a fixed template: RPTQ applies a block-level fixed permutation, and Atom fuses reordering around individual operator boundaries.
Neither provides a graph-level solution that propagates compatible channel bases across the full operator DAG, so permutation overhead resurfaces at block and layer boundaries outside the fused region.

\begin{figure}[t]
\centering
\includegraphics[width=\linewidth]{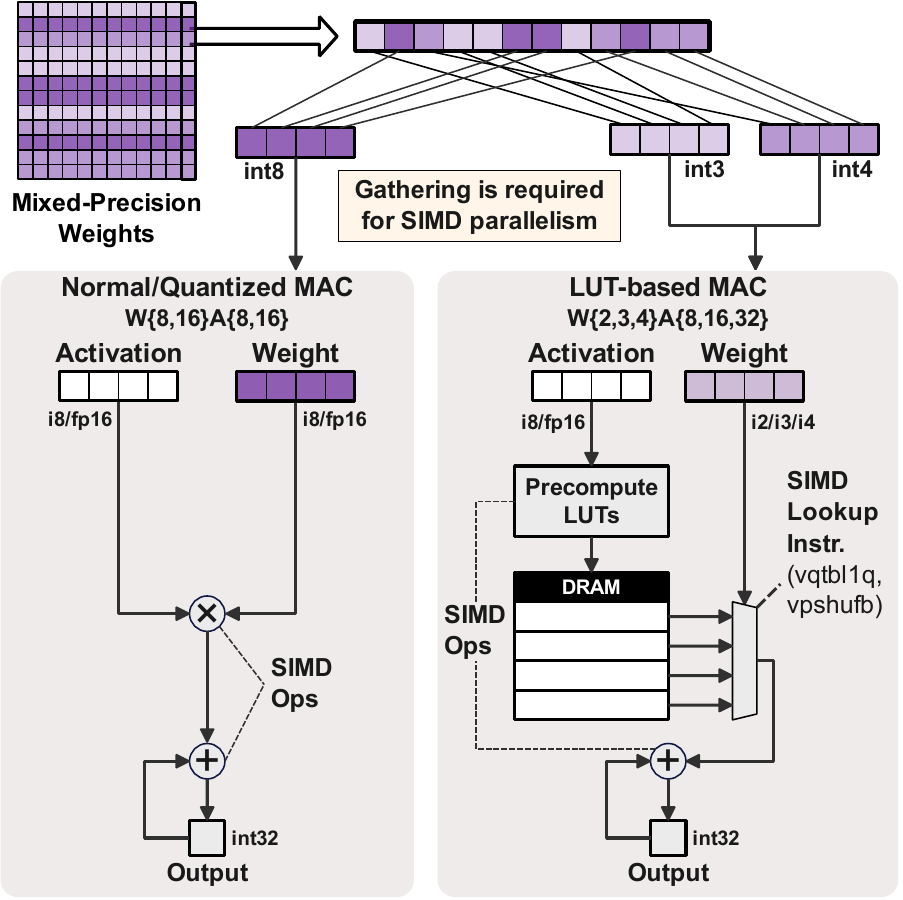}
\caption{
Kernel-side barrier for deployable fine-grained quantization on CPUs.
Conventional low-bit paths assume a uniform per-layer precision and map directly to SIMD MAC or LUT-based GEMM kernels with regular packing.
Channel-wise mixed precision introduces heterogeneous bit-widths across channels, creating two execution problems: layer-local blocks must still map to real low-bit kernel paths, and cross-operator permutations must not reappear as runtime data movement.
\design{} addresses this by clustering channels of equal precision, generating layer-specific kernels for the resulting blocks, and propagating compatible permutations ahead of time so LUT-based low-bit kernels and conventional SIMD paths coexist without sacrificing regularity.
}
\label{fig:kernel_req}
\end{figure}

Beyond palette and granularity limitations, deploying any fine-grained bit assignment on CPUs faces a second class of barriers on the execution side.
As sketched in \autoref{fig:kernel_req}, SIMD arithmetic assumes uniform data types across a vector: int8/int16 lanes for conventional MACs, or fixed sub-byte encodings for LUT-based low-bit GEMMs.
LUT-based MACs (e.g., 2--4\,bit weights implemented via table lookups with instructions such as \texttt{vpshufb} or \texttt{vqtbl1q\_u8}) can be extremely efficient~\cite{tmac, bitnet_cpp, tsar}, but only when all lanes share the same bit-width and packing scheme.
This creates a first barrier: even after bit assignment is decided, each layer must still be reorganized into bit-homogeneous blocks that map to actual kernel paths.
A second barrier appears across operators: if the required permutations are materialized between layers, extra unpacking, gather-like movement, and repacking quickly erase the benefit.
A practical CPU solution therefore needs both layer-local heterogeneous kernels and graph-level layout regularization.

\design{} is designed to resolve this split explicitly.
Section~\ref{sec:quantization} handles the \emph{accuracy-side} problem by assigning one of several discrete bit-widths per channel from a CPU-aligned \{2,3,4,8,16\}-bit palette to meet a user-specified fractional average bit target.
Section~\ref{sec:compiler} handles the \emph{execution-side} problem by clustering channels into bit-homogeneous blocks for layer-local kernels and propagating compatible permutations across operators to keep layout overhead off the runtime path.
This quantization/compiler co-design makes fine-grained budget control viable on CPUs: memory footprint and quality scale smoothly with the fractional bit budget, while runtime scaling follows the same budget parameter with modest overhead.
\section{Poly-Precision Quantization Method}
\label{sec:quantization}

\design{} performs per-channel, layer-wise bit allocation at \emph{compile time} under a user-specified \emph{fractional} average bit budget $B$ (e.g., $B=3.7$\,bits/weight).
Each input channel of each linear layer is first assigned a bit-width from $\{2,3,4,8,16\}$ by an idealized waterfilling step, and a later ISA-aware snap stage regularizes these counts to backend-friendly SIMD/LUT quanta.
This yields intra-layer mixed-precision weights that closely match a given memory budget while preserving accuracy and exposing a regular execution pattern to the CPU kernels.

\begin{figure}[t]
\centering
\includegraphics[width=\linewidth]{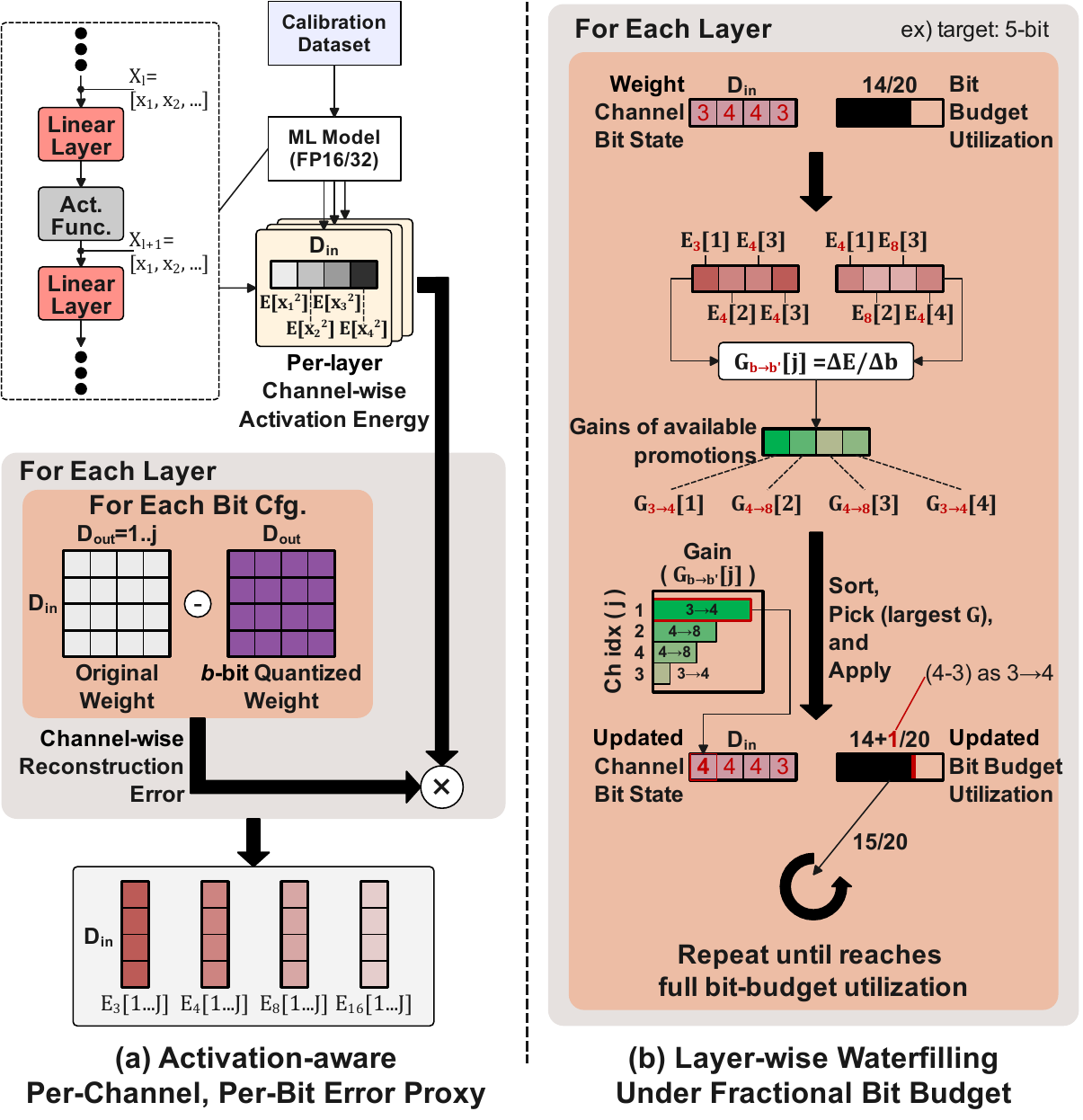}
\caption{
PolyQ quantization pipeline at the layer level.
(a) From a small calibration dataset, we run the FP16/FP32 model to collect per-layer, channel-wise activation energies $\mathbb{E}[x_j^2]$ and, for each linear layer and candidate bit-width $b \in \{2,3,4,8,16\}$, compute an activation-aware per-channel reconstruction error $E_b[j]$ over input channels ($D_{\mathrm{in}}$).
(b) For each layer, we treat bit allocation over input channels as a waterfilling problem under a fractional average bit budget: starting from an initial bit state (e.g., all 2-bit), we form gains $G_{b\to b'}[j] = \Delta E / \Delta b$ for all possible promotions, sort and apply the highest-gain feasible promotions while respecting the bit budget, and repeat until the residual budget cannot be used by another discrete promotion, yielding a mixed-precision bit state per input channel.
}
\label{fig:quantization}
\end{figure}

\subsection{Activation-aware per-channel error proxy}

We use an activation-aware error proxy per input channel, summarized in \autoref{fig:quantization}(a): weight reconstruction error is weighted by the empirical activation energy.
Consider a linear layer $\ell$ with weight matrix $W^{(\ell)} \in \mathbb{R}^{d_{\text{in}}^{(\ell)} \times d_{\text{out}}^{(\ell)}}$ and a per-channel scale vector $s^{(\ell)} \in \mathbb{R}^{d_{\text{in}}^{(\ell)}}_{>0}$ obtained from calibration.
For a candidate bit-width $b \in \{2,3,4,8,16\}$ and group size $g_b$, we form a scaled-and-quantized weight
\begin{equation}
  \tilde{W}^{(\ell,b)}
  \;=\;
  \operatorname{diag}\big((s^{(\ell)})^{-1}\big) \,
  Q_b\!\big(\operatorname{diag}(s^{(\ell)}) W^{(\ell)}\big)
\end{equation}
where $Q_b(\cdot)$ is a symmetric group-wise uniform quantizer with $b$-bit weights using the backend-compatible packing format for that precision.

We approximate the contribution of input channel $j$ in layer $\ell$ to the squared output error by
\begin{equation}
  E_b^{(\ell)}[j]
  \;\approx\;
  \underbrace{\big\| \tilde{w}^{(\ell,b)}_j - w^{(\ell)}_j \big\|_2^2}_{\text{weight reconstruction error}}
  \cdot
  \underbrace{\mathbb{E}\!\left[x_j^2\right]}_{\text{activation energy}},
  \label{eq:channel-error}
\end{equation}
where $w^{(\ell)}_j$ and $\tilde{w}^{(\ell,b)}_j$ are the $j$-th input-channel rows of $W^{(\ell)}$ and $\tilde{W}^{(\ell,b)}$, and $\mathbb{E}[x_j^2]$ is the second moment of the $j$-th input activation collected on a small calibration set.
Channels whose activations have large $\mathbb{E}[x_j^2]$ are more salient, and their quantization error is penalized more heavily.
For $b=16$ we treat the channel as unquantized and set $E_{16}^{(\ell)}[j]=0$.
All $|\{2,3,4,8,16\}|$ error values per channel are computed in a single pass over the weights and pre-computed activation statistics, making the proxy evaluation lightweight relative to the waterfilling step.

\subsection{Waterfilling under fractional bit budget}

\design{} treats each linear layer independently as a discrete resource allocation problem under a \emph{layer-wise} fractional bit budget, as illustrated in \autoref{fig:quantization}(b).
For layer $\ell$ with $d^{(\ell)}_{\text{in}}$ input channels, we wish to choose a bit-width $b_j^{(\ell)} \in \{2,3,4,8,16\}$ for each channel $j \in \{1,\dots,d^{(\ell)}_{\text{in}}\}$ such that
\begin{equation}
  \min_{\{b_j^{(\ell)}\}} \;
    \frac{1}{d^{(\ell)}_{\text{in}}} \sum_{j=1}^{d^{(\ell)}_{\text{in}}}
      E_{b_j^{(\ell)}}^{(\ell)}[j]
  \quad \text{s.t.} \quad
    \frac{1}{d^{(\ell)}_{\text{in}}} \sum_{j=1}^{d^{(\ell)}_{\text{in}}}
      b_j^{(\ell)} \;=\; B
  \label{eq:layer-alloc}
\end{equation}
where $B$ is the user-specified average bit budget shared by all layers (e.g., $B=3.7$).
In other words, each layer must, on average, use $B$ bits per input channel, but it is free to assign more bits to salient channels and fewer bits to unimportant ones.

We solve \autoref{eq:layer-alloc} by a greedy waterfilling.
Let $b_{\min}=2$ be the minimum bit-width.
We initialize all channels in layer $\ell$ to $b_{\min}$ consuming $b_{\min}\,d_{\text{in}}^{(\ell)}$ bits, and compute the remaining bit budget
\begin{equation}
  \Delta^{(\ell)} \;=\;
  \text{round}\!\big(B \cdot d_{\text{in}}^{(\ell)}\big)
  - b_{\min}\,d_{\text{in}}^{(\ell)}
\end{equation}
If $\Delta^{(\ell)} \le 0$, all channels stay at 2\,bit.
Otherwise, we repeatedly \emph{upgrade} channels by assigning extra bits where they reduce the error the most:

\noindent(1) For each channel $j$, we consider upgrading it from its current bit-width $b$ to the next higher supported value $b' \in \{2,3,4,8,16\}$ with $b' > b$ and compute the gain in error per additional bit
\begin{equation}
  G_{b \rightarrow b'}^{(\ell)}[j]
  \;=\;
  \frac{E_b^{(\ell)}[j] - E_{b'}^{(\ell)}[j]}{b' - b}
  \label{eq:gain}
\end{equation}

\noindent(2) We maintain a max-heap of all candidate upgrades across channels in layer $\ell$, keyed by $G_{b \rightarrow b'}^{(\ell)}[j]$.

\noindent(3) While $\Delta^{(\ell)} > 0$ and a feasible upgrade remains, we pop the candidate with the largest gain that still matches the current bit-width assignment and fits the remaining budget, apply the upgrade $b \rightarrow b'$, and decrease $\Delta^{(\ell)}$ by $(b' - b)$.
        After upgrading channel $j$, we insert its next possible upgrade (e.g., $4 \rightarrow 8$) into the heap.

This per-layer waterfilling algorithm starts from an aggressively quantized configuration and spends additional bits on the channels that yield the largest reduction in activation-weighted error per extra bit, until the per-layer budget $B$ is exhausted or no exact discrete promotion fits the residual budget.
Because each layer may have a different saliency pattern, the resulting 2/3/4/8/16-bit mix can vary across layers, even though the pre-snap assignment in each layer targets the same average bit constraint.

\subsection{ISA-aware SIMD-quanta matching}
The waterfilling solution above optimizes an \emph{ideal} per-channel bit map, but the execution back-end still requires backend-friendly block sizes.
Concretely, our current CPU path targets $W\{2,3,4\}A16$ LUT-based kernels together with $W\{8,16\}A16$ SIMD kernels, so quanta matching regularizes the weight-channel layout for these kernel families.
For a target ISA/back-end $t$, we therefore associate each supported bit-width $b$ with a channel quantum $q_b^{(t)}$ determined by the packing format and SIMD microkernel.
For example, AVX2 int8 kernels naturally consume channels in multiples of 32, while LUT-based low-bit kernels impose their own per-bit quanta;
the same abstraction can be retargeted to ARM NEON/SVE or other CPU kernels by substituting their backend-specific $\{q_b^{(t)}\}$.

For layer $\ell$, let $c_b^{(\ell)}$ be the number of channels assigned to bit-width $b$ by the waterfilling stage.
We first round these counts to a nearby realizable target
\vspace{-1mm} 
\begin{equation}
  \hat{c}_b^{(\ell)}
  \;\approx\;
  \operatorname{round}\!\left(\frac{c_b^{(\ell)}}{q_b^{(t)}}\right) q_b^{(t)},
  \qquad
  \sum_b \hat{c}_b^{(\ell)} = d_{\text{in}}^{(\ell)},
\vspace{-1mm} 
\end{equation}
using small residual corrections to preserve the total number of channels.
The rounded counts define target block sizes for code generation, but leave channel membership to the subsequent error-aware reassignment step.

We apply a second error-aware reassignment step, using equation \eqref{eq:gain}.
When a higher-precision block needs more channels, we promote the channels with the largest activation-aware gain and when a block must shrink, we demote the channels with the smallest loss under the same criterion.
In practice we restrict moves to neighboring bit levels and iterate until the per-bit counts match $\{\hat{c}_b^{(\ell)}\}$.
Thus the snap stage is not a naive padding rule over the raw waterfilling map: it explicitly re-optimizes the channel membership of each bit block so the final layout is hardware-realizable with minimal proxy-error increase.

After quanta matching, the compiled layer realizes
\vspace{-1mm} 
\begin{equation}
  B_{\mathrm{real}}^{(\ell)}
  \;=\;
  \frac{1}{d_{\text{in}}^{(\ell)}} \sum_{j=1}^{d_{\text{in}}^{(\ell)}} \tilde{b}_j^{(\ell)},
\vspace{-1mm} 
\end{equation}
By combining nearby-count rounding with error-aware channel reassignment, \design{} keeps $B_{\mathrm{real}}^{(\ell)}$ close to the requested $B$ while matching discrete kernel quanta.

\subsection{Bit-specific activation-aware scaling}

After the per-layer bit maps $\{\text{bits\_map}_\ell\}$ are fixed by the waterfilling and quanta matching steps, we apply a bit-specific activation-aware scaling.
Let $m^{(\ell)} \in \mathbb{R}^{d_{\text{in}}^{(\ell)}}_{>0}$ denote an activation-based saliency metric for layer $\ell$ (we use normalized second moments or mean absolute activations on the calibration set).
For each bit $b \in \{2,3,4,8,16\}$, we share a single bit-specific exponent $\alpha_b \in [0,1]$ across all input channels in layer $\ell$ that use bit-width $b$, and define
\vspace{-1mm} 
\begin{equation}
  s^{(\ell)}_j
  \;=\;
  \big(m^{(\ell)}_j\big)^{\alpha_b}
  \quad\text{for } j \in \mathcal{I}^{(\ell)}_b
  \;=\;
  \{ j : \text{bits\_map}_\ell[j] = b \}.
\vspace{-1mm} 
\end{equation}
We choose $\alpha_b$ by a small grid search that minimizes the average proxy error
$\frac{1}{|\mathcal{I}^{(\ell)}_b|}\sum_{j \in \mathcal{I}^{(\ell)}_b} E_b^{(\ell)}[j]$.
This yields per-channel scales that are still described by a single power law per bit-width, but adapted to the poly-precision bit layout.
Because each bit group in a layer shares a single exponent $\alpha_b$, the resulting scales are cheap to store and easy to fold into the packed weight layout that our kernel compiler consumes.

In our implementation, the final weights are obtained by applying standard group-wise symmetric quantization at the assigned bit-width per input channel using these scales, and then using the resulting packed weights in our CPU kernels.
\section{Model Compiler Architecture}
\label{sec:compiler}

Given the per-channel bit maps from \autoref{sec:quantization}, the \design{} compiler transforms irregular channel-wise bit assignments into a static profile 
of permuted, packed weights and layer-specific kernels, executable without 
run-time reconfiguration. \autoref{fig:compiler} summarizes the key steps.

\subsection{Channel permutation and clustering}
\begin{figure}[t]
\centering
\includegraphics[width=\linewidth]{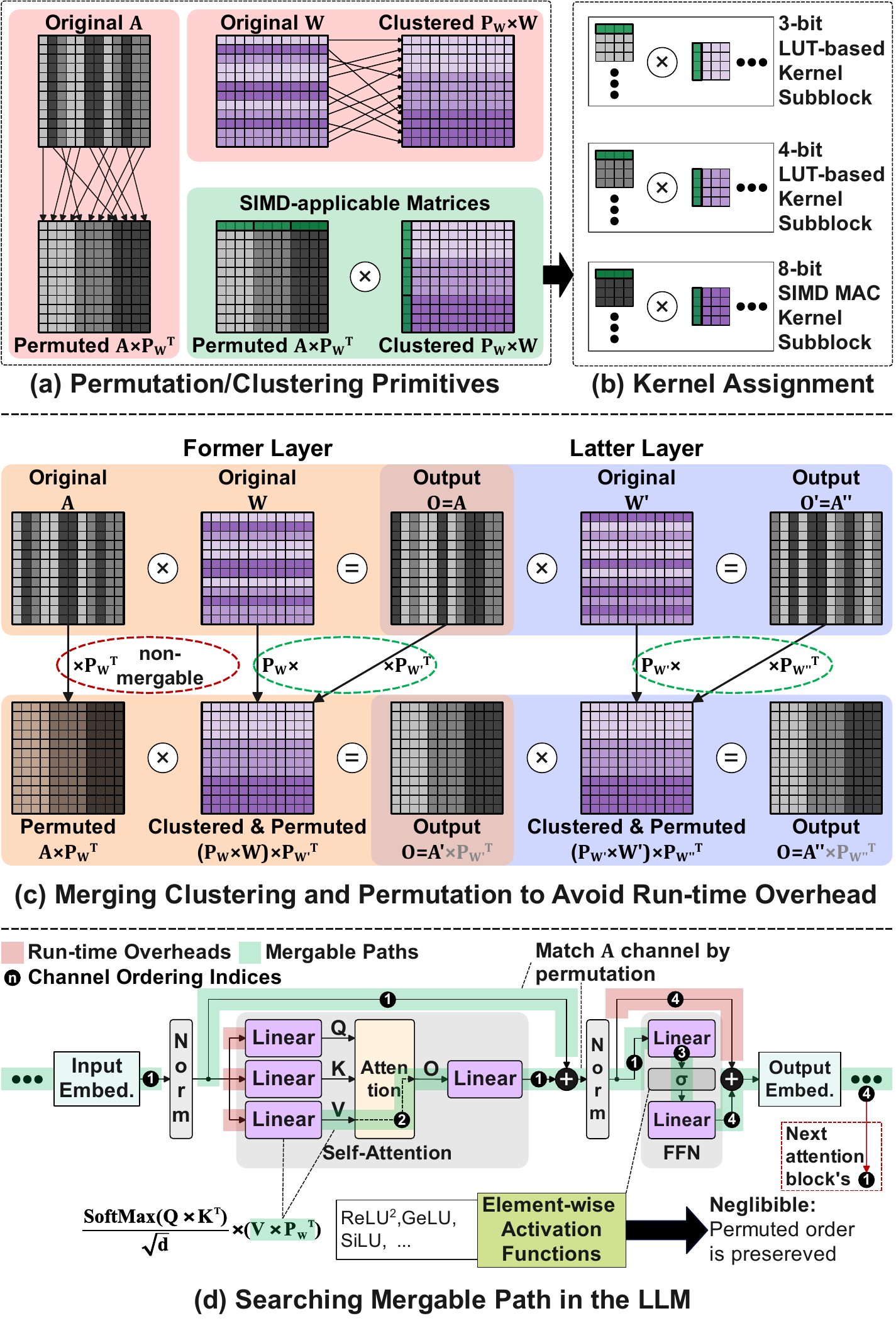}
\caption{
PolyQ model compiler architecture.
(a) A channel permutation factors $O=A\times W=(A\times P_W^\top)\times(P_W\times W)$, clustering equal-bit input rows without changing the linear output.
(b) Clustered input rows are tiled into $W\{2,3,4\}A16$ LUT and $W\{8,16\}A16$ SIMD subblocks.
(c) Across adjacent linears, a downstream activation shuffle is absorbed into compile-time output/input weight permutations.
(d) A static operator-DAG pass propagates channel bases through basis-preserving nodes and inserts shuffles only at incompatible fan-out, fan-in, or path-boundary edges.
}
\label{fig:compiler}
\end{figure}

For each linear layer with input activations $A$ and weight matrix $W$, we introduce a channel clustering and permutation matrix $P_W$ that reorders input channels so that input channels with the same bit-width become contiguous, as shown in \autoref{fig:compiler}(a).
Conceptually, we can factor the linear as
\begin{equation}
  O = A \times W = (A \times P_W^\top)\times(P_W \times W),
\end{equation}
where $P_W \times W$ is a bit-clustered weight with input-channel rows made contiguous and $A \times P_W^\top$ is the correspondingly permuted activation.
At compile time we materialize only the clustered weight $P_W \times W$; the permutation matrix itself is never stored at runtime.
The same compile-time planner propagates the corresponding channel permutations across consecutive linears, explained in Sections \ref{sec:merge} and \ref{sec:plan}, so mergeable layers run directly with $A$ interpreted in this permuted basis,
and only incompatible graph boundaries pay for an explicit activation shuffle.

\subsection{Bit blocks and kernel assignment}
After the quanta-matching step in \autoref{sec:quantization}, each clustered weight matrix decomposes into a small number of bit-homogeneous input-channel blocks (2/3/4/8/16\,bit).
The compiler assigns a kernel per block, shown in \autoref{fig:compiler}(b).
Low-bit blocks use $W\{2,3,4\}A16$ LUT-based GEMM/GEMV kernels, while higher-precision blocks use $W\{8,16\}A16$ OpenBLAS SIMD GEMM~\cite{openblas}, since their larger group sizes make table-lookup less efficient than direct MAC operations.
This produces a fixed sequence of block kernels per linear layer, written as C++ functions.

\subsection{Merging permutations across layers}\label{sec:merge}
Applying $P_W^\top$ on activations would introduce run-time overhead.
The compiler therefore uses a simple merge rule: a downstream activation permutation can be eliminated if the upstream producer can emit its output in that same channel basis.
For two consecutive linears with $O = A \times W$ and $O' = O \times W'$, suppose the second layer wants its input in basis $P$.
Instead of materializing $O \times P^\top$ at run time, the compiler rewrites
\begin{equation}
O'
=
(O \times P^\top) \times (P \times W')
=
A \times (W \times P^\top) \times (P \times W') .
\end{equation}
Thus the downstream activation permutation is replaced by compile-time permutations of the upstream output-channel columns and downstream input-channel rows.
The intermediate activation is produced directly in the basis required by the downstream layer.
Because non-linear activation functions such as ReLU/GELU operate independently per channel, they commute with channel permutations and forward the same basis unchanged.
This same rule lets the last linear in one region produce the basis required by the first linear in the next region, so permutation merging is not limited to boundaries inside a single Transformer block.

\subsection{Graph-based permutation planning}\label{sec:plan}
Rather than hard-coding attention templates, the compiler treats the model as a static operator directed acyclic graph (DAG) whose edges are activation tensors and whose nodes are linears, normalizations, element-wise activations, residual adds, attention operators, and reshapes.
Each linear node has a preferred input-channel permutation induced by its snapped bit blocks, while each edge carries a current channel basis.
The compiler then assigns channel bases to DAG edges and forms maximal regions in which the merge rule above can be applied without materializing an activation shuffle.

Equivalently, planning is a finite graph-labeling problem: assign a channel-basis label $\pi_e$ to each activation edge $e$, with labels drawn from the permutations requested by neighboring mixed-precision linears, and reduce the number of edges that require an explicit shuffle.
The local compatibility rules define the cost.
Basis-preserving operators such as normalization and element-wise activations require identical input/output labels, linears can absorb the input label into their packed weight and choose an output label by permuting output channels at compile time, and fan-in/fan-out/path-boundary nodes are zero-cost only when their incident labels agree.
Our implementation uses topological propagation with local conflict resolution: keep the label with the largest downstream reuse and insert shuffles only on the remaining boundary edges.
This discovers mergeable paths beyond a single Transformer block, including cases where the final FFN linear's output basis matches the next attention block's first projection.

\section{Evaluation}
\label{sec:eval}
We evaluate \design{} along four questions: whether a CPU-aligned palette improves quality over prior methods at matched budgets, whether fractional budgets survive ISA-aware realization and map to practical memory footprints, whether compiler layout regularization reduces activation reorder traffic, and whether CPU execution remains predictable.

\subsection{Experimental Setup}
\noindent\textbf{Quantization baselines.}
We compare \design{} with AWQ, GPTQ, Slim-LLM, and AMQ~\cite{awq,gptq,slimllm,amq} using group size 128 and FP16 activations, calibrated on 128 randomly sampled WikiText-2 sequences.
Baselines are evaluated at their supported integer-only or fractional targets; \design{} additionally sweeps fractional budgets in 0.1-bit steps.

\noindent\textbf{Models and quality metrics.}
We evaluate Falcon-H1-3B~\cite{falconh1}, Llama2-13B~\cite{llama2}, and Qwen3-32B~\cite{qwen3} on WikiText-2~\cite{wikitext} perplexity~\cite{awq,gptq} and capped downstream accuracy on MMLU~\cite{mmlu}, WinoGrande~\cite{winogrande}, ARC-Easy/Challenge~\cite{arc}, and HellaSwag~\cite{hellaswag}, with question sets fixed across methods and bit budgets. Downstream averages are computed over these five tasks.

\noindent\textbf{CPU execution setup.}
We measure end-to-end execution on Ryzen 9 9950X, Ryzen 7 7840U, and Intel N250 (workstation, laptop, and mobile, 16/8/4 cores) running Ubuntu 24.04. Energy/token is measured via RAPL \cite{rapl} OS module. \design{} dispatches $W\{2,3,4\}A16$ blocks to T-MAC~\cite{tmac} and $W\{8,16\}A16$ blocks to OpenBLAS SIMD GEMM~\cite{openblas}. Runtime comparison uses uniformly quantized AWQ weights on the same T-MAC path at 2-, 3-, and 4-bit.

\subsection{Accuracy and Low-Bit Budget Recovery}
\label{sec:eval-accuracy}

\begin{figure}[t]
\centering
\includegraphics[width=\linewidth]{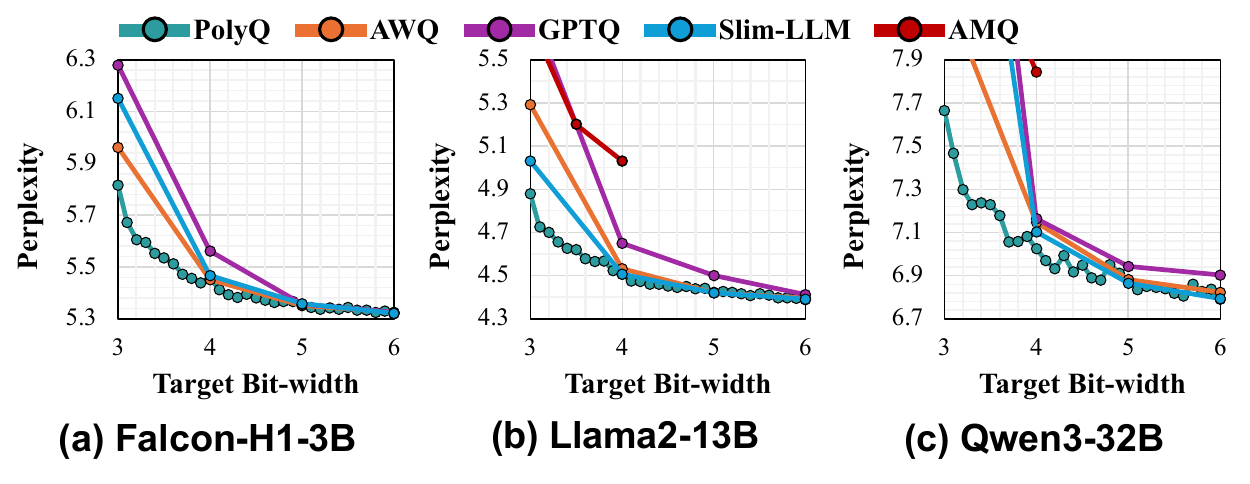}
\caption{
Perplexity vs.\ target average bit budget for the three models.
\design{} sweeps the fractional bit budget from 3.0\,b to 6.0\,b in 0.1-bit steps, 
while AWQ, GPTQ, and Slim-LLM are evaluated at supported integer targets and 
AMQ at its 3\,b, 3.5\,b, 4\,b Pareto-efficient point within this range.
}
\label{fig:ppl-bits}
\end{figure}

\noindent\textbf{Fractional-bit quality analysis.}
\autoref{fig:ppl-bits} shows perplexity as a function of the target average bit budget.
\design{} exposes a dense fractional ladder from 3.0\,b to 6.0\,b in 0.1-bit steps, while AWQ, GPTQ, and Slim-LLM provide coarse integer reference points and AMQ presents few Pareto-efficient points.
Across all three models, increasing $B$ improves perplexity smoothly with only small local fluctuations, and higher-precision \design{} points converge to the corresponding integer-baseline quality.
At 5--6\,b, all methods produce similar perplexity, consistent with channels naturally settling near their target precision regardless of palette choice;
Slim-LLM's marginal advantage at these budgets is consistent with its b$\pm$1 palette providing finer intermediate steps between 4 and 8\,bit, though the gap is within 0.01--0.03 perplexity.
The meaningful differentiation occurs at tight budgets: at 3\,b, \design{} outperforms AWQ by 2.4--7.8\%, GPTQ by 7.4--28.7\%, Slim-LLM by 3.0--23.4\%, and AMQ by 14.1--32.1\%, where channel-level granularity and a CPU-aligned palette spanning 2--16-bit matter most.
We use 3.0--6.0\,b as the stable operating window and analyze the 2.x low-bit floor separately below.

\noindent\textbf{Standard 3-bit operating point.}
\autoref{tab:3bit} details the bit distribution at the 3-bit target. Most channels remain in the 2--4\,b range, while only small fractions are promoted to 8/16\,b where the activation-aware proxy marks them as salient, keeping the realized average within 0.016\,b of the target.

\begin{table}[t]
\centering
\small
\caption{
\design{} perplexity and bit distribution at a 3-bit target.
}
\label{tab:3bit}
\resizebox{\linewidth}{!}{%
\begin{tabular}{lccccc|ccccc}
\toprule
 & \multicolumn{5}{c|}{Perplexity ($\downarrow$ is better)} & \multicolumn{5}{c}{\design{} bit dist. (\%) at 3\,b} \\
Model & AWQ & GPTQ & Slim-LLM & AMQ & \design{} & 2b & 3b & 4b & 8b & 16b \\ \midrule
Falcon-H1-3B & 5.96 & 6.28 & 6.15 & --$^\dagger$ & \textbf{5.82} & 6.61 & 86.23 & 6.92 & 0.24 & 0.00 \\
Llama2-13B   & 5.29 & 5.75 & 5.03 & 5.68 & \textbf{4.88} & 4.08 & 92.33 & 3.42 & 0.16 & 0.01 \\
Qwen3-32B    & 8.23 & 10.75 & 10.00 & 11.29 & \textbf{7.66} & 11.93 & 77.82 & 9.68 & 0.56 & 0.01 \\
\bottomrule
\multicolumn{11}{l}{$^\dagger$AMQ currently does not support Falcon-H1-3B's hybrid attention architecture.} \\
\end{tabular}}
\end{table}

\noindent\textbf{Extreme low-bit transition.}
\autoref{tab:min-bits} highlights why fractional precision is most valuable near the low-bit floor.
At this end of the design space, integer precision steps are coarse: moving from W2 to W3 increases weight storage by 50\%.
\design{} identifies model-specific recovery points $B_{min}$ (defined in \autoref{tab:min-bits}), the first fractional operating point past the W2 collapse where perplexity recovers to a usable regime, avoiding both collapsed W2-like quality and overprovisioned W3 memory.
These $B_{min}$ result in 1.47\,GiB, 4.98\,GiB, and 11.29\,GiB estimated peak footprints for Falcon-H1-3B, Llama2-13B, and Qwen3-32B, respectively.
Compared with the same-backend W3 footprint, this leaves 17.1\%, 11.6\%, and 21.8\% less peak memory at the first recovered operating point, which is memory that cannot be reclaimed in an integer-only deployment.

\begin{table}[t]
\centering
\small
\caption{
Low-bit recovery threshold and cost of uniform W3 fallback.
$B_\text{min}$ is the first swept point after the collapse-to-recovery transition; W3 overhead is computed as $3/B_\text{min}-1$.
}
\label{tab:min-bits}
\resizebox{\linewidth}{!}{%
\begin{tabular}{lccc}
\toprule
Model & \makecell{Collapse point ($B$/PPL)} & \makecell{Recovery point ($B_\text{min}$/PPL)} & \makecell{W3 overhead} \\
\midrule
Falcon-H1-3B & 2.0 / $2.0{\times}10^7$ & 2.1 / 8.67  & 42.9\% \\
Llama2-13B   & 2.5 / 44.78             & 2.6 / 5.96  & 15.4\% \\
Qwen3-32B    & 2.1 / $1.4{\times}10^5$ & 2.2 / 11.39 & 36.4\% \\
\bottomrule
\end{tabular}}
\end{table}

\begin{figure}[t]
\centering
\includegraphics[width=\linewidth]{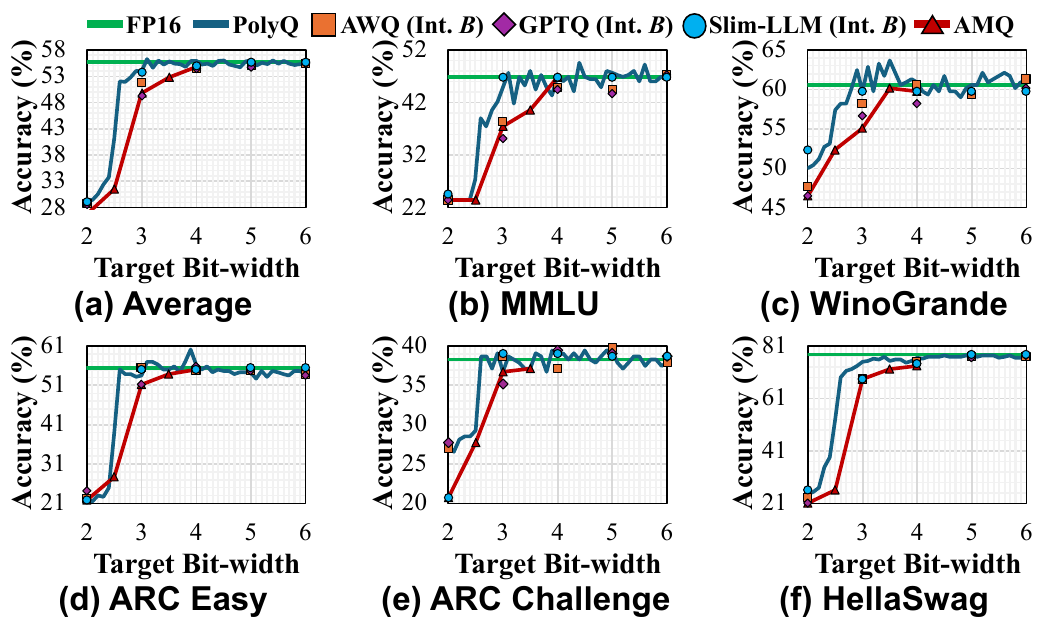}
\caption{
Capped downstream multiple-choice accuracy vs.\ target bit budget.
\design{} is plotted as a fractional-budget curve; AWQ, GPTQ, and Slim-LLM are integer-budget markers, and AMQ is plotted at its evaluated Pareto points.
}
\label{fig:graph_downstream}
\end{figure}

\noindent\textbf{Downstream task behavior.}
\autoref{fig:graph_downstream} shows that, in this capped evaluation, downstream results follow the same low-bit recovery region as perplexity.
We detail Llama2-13B as it shows the sharpest low-bit recovery transition; Qwen3-32B and Falcon-H1-3B follow with briefer summaries.
For Llama2-13B, the five-task average rises from 41.3 at $B{=}2.5$ to 52.0 at $B{=}2.6$, matching the perplexity drop from 44.78 to 5.96.
This recovery is broad rather than driven by a single benchmark: MMLU, ARC-Easy, ARC-Challenge, and HellaSwag improve by 11.7, 16.0, 9.4, and 16.0 percentage points, while WinoGrande remains stable.
By $B{=}2.9$, Llama2-13B reaches a 54.1 five-task average, above AMQ-3 at 49.8, AWQ-3 at 51.8, and Slim-LLM-3 at 53.8, while 
still using a sub-3-bit target.
For Qwen3-32B, the average rises from 32.0 at $B{=}2.1$ to 38.8 at $B{=}2.2$, and reaches 44.1 at $B{=}3.0$, slightly above the AWQ-3 marker at 43.2.
The Qwen3-32B transition is more uneven across tasks, but the large gains on HellaSwag and WinoGrande show the same recovery of task utility once the collapsed low-bit region is crossed.
Falcon-H1-3B stays in the same accuracy band as FP16 and integer baselines across 4--5\,b, and downstream curves across all models show small non-monotonic variations after the recovery point, consistent with the sensitivity of discrete benchmark metrics to evaluation scale~\cite{fluctuation}.

\subsection{Budget Realization and Peak Footprint}
\label{sec:eval-budget}

\begin{figure}[t]
\centering
\includegraphics[width=\linewidth]{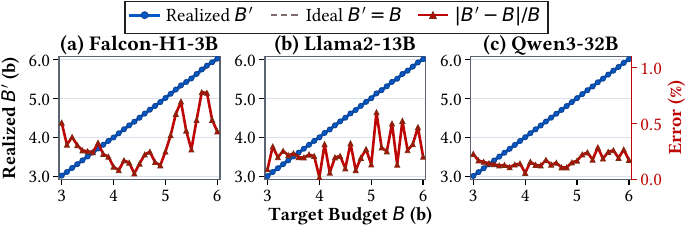}
\caption{
Budget realization after ISA-aware quanta matching from 3.0 to 6.0 target bits.
Markers show realized average precision $B'$ after snapping to backend-friendly SIMD/LUT quanta, the dashed line is the ideal $B'{=}B$ target, and the secondary axis reports relative mismatch.
}
\label{fig:quanta-match}
\end{figure}

\noindent\textbf{ISA/Kernel-aware quanta matching.}
\design{} preserves the requested target average bit budget after snapping mixed-precision channel counts to SIMD/LUT-friendly quanta.
\autoref{fig:quanta-match} confirms this holds across the 3.0--6.0\,b sweep: the maximum observed mismatch is only 0.045\,b on Falcon-H1-3B, 0.031\,b on 
Llama2-13B, and 0.016\,b on Qwen3-32B, staying within 0.016\,b and 0.005\,b at $B{=}3$ and $B{=}4$, respectively.
The snap step also keeps residual padding modest, reducing raw-map padding---unused tile slots before quanta matching---from 1.30--5.77\% to 0.46--1.68\%.

\begin{figure}[t]
\centering
\includegraphics[width=\linewidth]{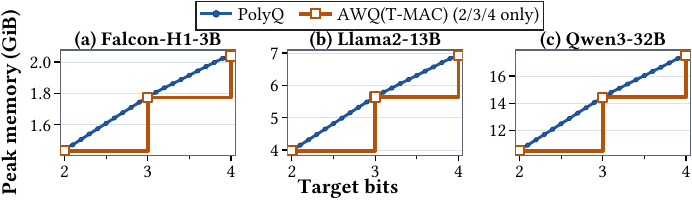}
\caption{
Peak memory footprint on the three models under fixed batch$\,{=}\,1$.
The footprint model combines quantized storage with KV cache, live activations, backend scratch, and layout metadata.
\design{} exposes 0.1-bit control from 2.0 to 4.0 bits, while AWQ(T-MAC) moves only in coarse 2/3/4-bit steps.
}
\label{fig:budget-fit}
\end{figure}

\noindent\textbf{Peak memory footprint.}
\autoref{fig:budget-fit} translates the fractional budget into a fixed-regime peak-footprint study under fixed batch$\,{=}\,1$, modeling peak memory as the 
maximum of a 256-token prefill and a 1-token decode working set with 256 cached tokens. Unlike prior work reporting weight-only theoretical savings, our model 
accounts for all live buffers: quantized storage, FP16 residue, KV cache, live activations, backend scratch, layout metadata, and one reorder buffer.
At matching integer budgets, \design{} remains within 1.1\% of the same-backend AWQ(T-MAC) footprint across all three models, confirming that per-channel bit maps and permutation metadata add negligible overhead relative to the quantized weights.
The key benefit is the dense set of intermediate footprints: instead of choosing only W2/W3/W4, a deployment can select budgets such as 2.6\,b or 3.4\,b and obtain corresponding peak-memory points.
For Llama2-13B at $B{=}2.6$, storage accounts for 4.75\,GiB of the 4.98\,GiB peak footprint, while layout metadata plus the reorder buffer contribute only 
0.006\,GiB, confirming the fractional footprint is controlled primarily by the weight budget, not mixed-precision bookkeeping.

\begin{figure*}[t]
\centering
\includegraphics[width=\linewidth]{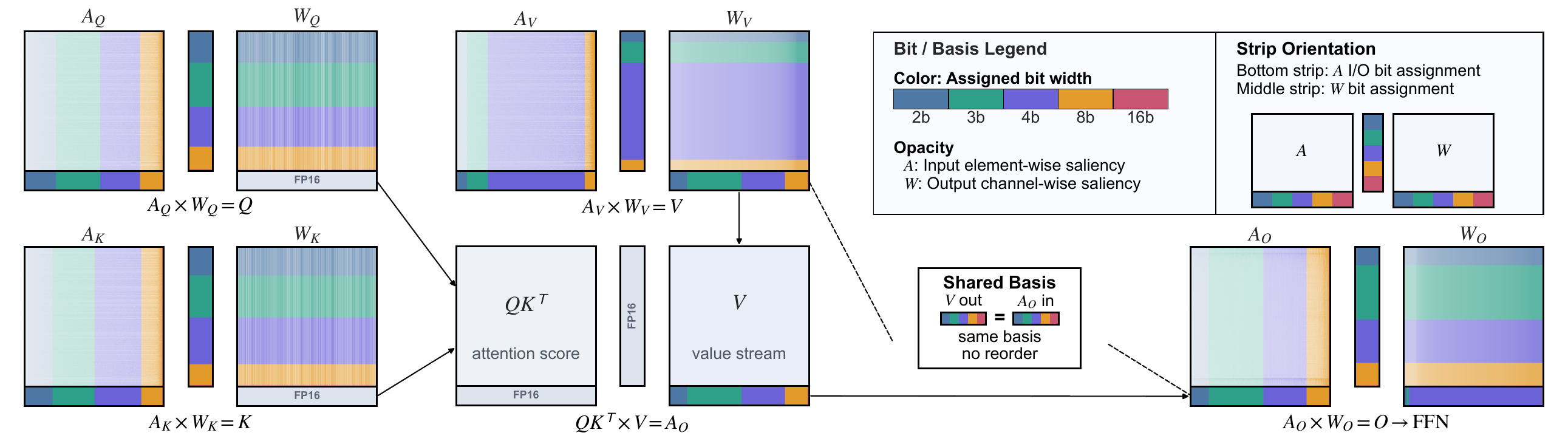}
\caption{
Qualitative compiler layout example from the first transformer block of Llama2-13B at $B{=}4$.
The view covers the Q/K/V projections and output projection; colors indicate assigned weight precisions after budget allocation and quanta matching.
PolyQ groups irregular channel assignments into contiguous same-bit groups and propagates a compatible channel order across the block, reducing explicit reorder sites before code generation.
}
\label{fig:graph-permute-cluster}
\end{figure*}

\subsection{Compiler Layout Regularization}
\label{sec:eval-compiler}

\noindent\textbf{Representative clustering outcome.}
\autoref{fig:graph-permute-cluster} visualizes the compiler transformation inside the first transformer block of Llama2-13B.
Each matrix is shown with bit-width strips and opacity maps: colors show assigned precision after budget allocation, while opacity marks input- or output-channel saliency.
Without compilation, high-saliency channels appear as scattered islands requiring frequent activation shuffles between operators.
\design{} turns this irregular map into a compile-time layout problem: it permutes channels so same-bit groups become contiguous, applies quanta matching so those groups fit kernel tile sizes, and propagates one compatible channel order across the Q/K/V projections and output projection.
The fixed-size tile step leaves a few unused tile slots when a group size is not divisible by the tile size, but these slots add only 0.58\% to the packed representation in this Llama2-13B trace.
The resulting layout absorbs local activation reorders into weight permutations and emits 30\% and 35\% fewer transitions between adjacent bit-specific packed blocks at $B{=}3$ and $B{=}4$, respectively.

\begin{figure}[t]
\centering
\includegraphics[width=\linewidth]{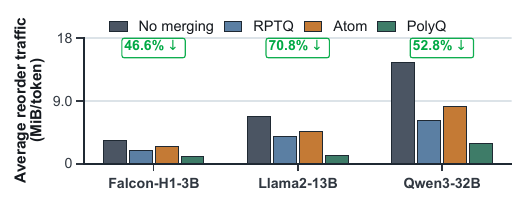}
\caption{
Modeled activation reorder traffic on the runtime path, averaged over $B{=}3.0$--$4.0$.
Bars count A16 read+write traffic for explicit activation reorders.
Top labels report \design{}'s average reduction relative to the best Atom/RPTQ-style fixed policy~\cite{atom,rptq}.
}
\label{fig:compiler-evidence}
\end{figure}

\noindent\textbf{Reorder-traffic reduction.}
\autoref{fig:compiler-evidence} isolates layout policy using a per-token activation-traffic model that counts A16 read+write bytes for explicit activation reorders, since end-to-end runtime would otherwise conflate reorder planning with quantizer, packing, and scheduler differences.
RPTQ-style planning represents fixed template-based reorder fusion inside a transformer block, Atom-style planning represents local fusion around mixed-precision operator boundaries, and \design{} applies DAG-level propagation on the entire model.
Without merging, explicit activation reorders account for 3.28, 6.80, and 14.50\,MiB per token on Falcon-H1-3B, Llama2-13B, and Qwen3-32B, respectively.
The RPTQ-style template gives 1.88, 3.91, and 6.25\,MiB per token, while the Atom-style local policy gives 2.53, 4.69, and 8.25\,MiB per token.
\design{} further reduces this traffic to 1.00, 1.14, and 2.95\,MiB per token, corresponding to 46.6\%, 70.8\%, and 52.8\% less than the best prior-style policy.
Thus, the compiler contribution is not a different bit allocation; it is the removal of residual runtime layout traffic that fixed local or template-based fusion leaves behind.

\begin{figure}[t]
\centering
\includegraphics[width=\linewidth]{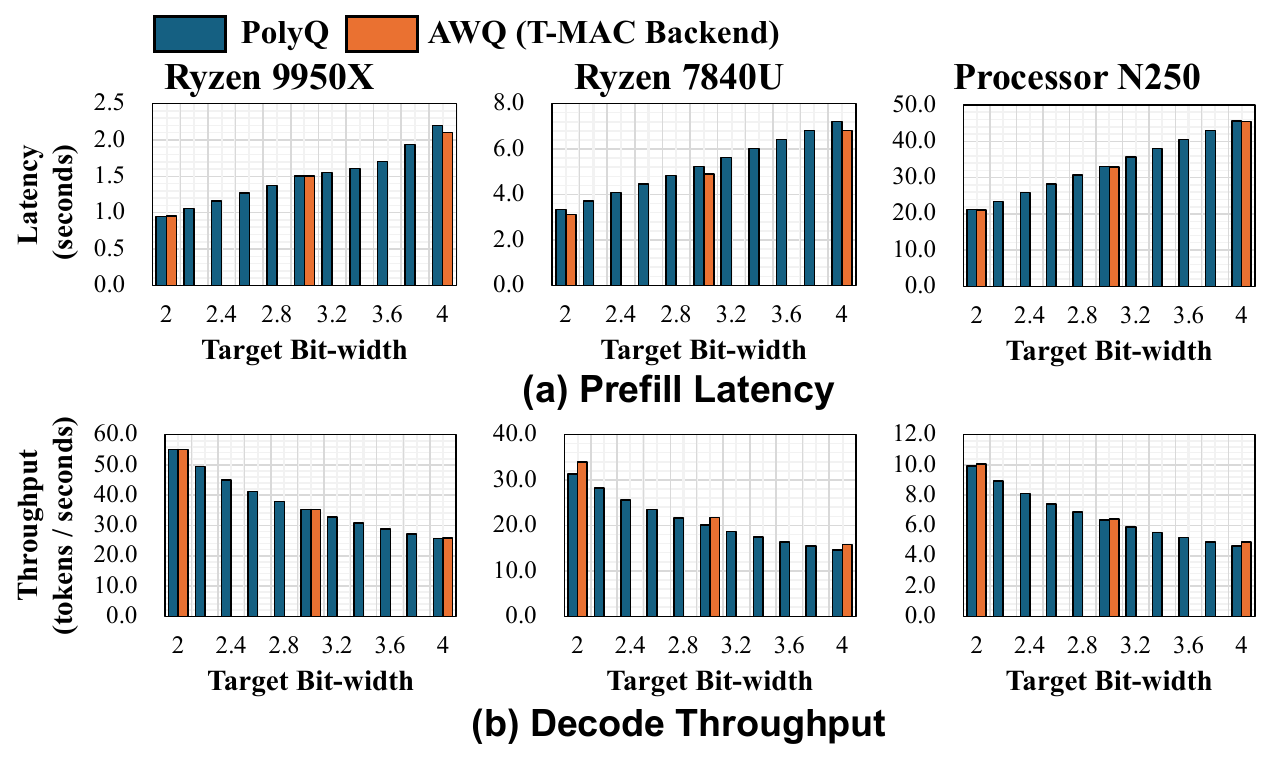}
\caption{
CPU system behavior of \design{} on three CPU targets as a function of target average bit budget $B$ for Llama2-13B.
(a) Prefill latency for a 256-token prompt.
(b) Batch$\,{=}\,1$ decode throughput on the same platforms.
\design{} closely tracks the AWQ baseline executed on the same T-MAC low-bit path, while both latency and throughput scale predictably with $B$.
}
\label{fig:system}
\end{figure}

\subsection{CPU Throughput and Energy}
\label{sec:eval-system}
\autoref{fig:system} reports Llama2-13B prefill latency and batch$\,{=}\,1$ decode throughput, comparing \design{} with AWQ on the same T-MAC low-bit backend for apples-to-apples timing. Slim-LLM, RPTQ, and Atom implement GPU-only execution backends and have no CPU inference path, precluding direct runtime comparison on T-MAC.
Across platforms and bit budgets, \design{} stays close to this integer-budget reference.
For the integer-budget reference points between $B{=}2$ and $B{=}4$, prefill latency stays within 4.7\% of AWQ(T-MAC) on Ryzen 9950X, within 5.8--7.3\% on Ryzen 7840U, and within 0.1--0.4\% on Intel N250.
Decode throughput shows the same modest gap: the shortfall is about 0.2\% on Ryzen 9950X, 8.1--8.2\% on Ryzen 7840U, and 1.2--6.1\% on N250, preserving the expected inverse scaling with bit budget.
Because both systems share the dominant LUT path for 2/3/4-bit work, runtime mainly tracks quantized weight traffic; the remaining gap comes from residual layout/control overhead and sparse OpenBLAS dispatches for \design{}'s higher-precision blocks.

\noindent\textbf{Scaling with fractional bits.}
Across the three CPUs, increasing $B$ from 2 to 3 raises \design{} prefill latency by 1.56--1.58$\times$ (0.95--1.51\,s on Ryzen 9950X, 3.35--5.24\,s on Ryzen 7840U, and 21.17--33.09\,s on Intel N250), close to the ideal 1.50$\times$ increase in weight traffic.
Increasing $B$ from 3 to 4 raises latency by another 1.38--1.46$\times$, close to the ideal 4/3 ratio.
Decode throughput mirrors this trend, dropping by 1.56$\times$ from $B{=}2$ to $B{=}3$ and by 1.37--1.38$\times$ from $B{=}3$ to $B{=}4$ across the same platforms.
Thus the same budget parameter that controls peak footprint in \autoref{fig:budget-fit} also gives a monotonic latency/throughput tradeoff, turning the budget sweep into a usable design curve rather than three coarse W2/W3/W4 choices.

\noindent\textbf{Energy per token.}
\autoref{tab:energy} shows that \design{} incurs only 0.8--1.9\% measured energy/token overhead relative to AWQ(T-MAC) at $B{=}3$ across all three platforms.
Since both systems share the dominant LUT-based kernels for 2/3/4-bit work, fractional poly-precision adds deployment flexibility at negligible energy cost.
The overhead is consistently modest across all three CPU classes --- workstation, laptop, and mobile --- suggesting that the mixed-precision bookkeeping cost scales with 
neither core count nor memory bandwidth.

\begin{table}[t]
\centering
\small
\caption{
Energy per token for Llama2-13B at $B{=}3$ from CPU execution.
}
\label{tab:energy}
\resizebox{\linewidth}{!}{%
\begin{tabular}{lccc}
\toprule
Platform & AWQ+T-MAC [J/token] & \design{} [J/token] & Overhead [\%] \\
\midrule
Ryzen 9 9950X & 6.03 & 6.10 & 1.13\% \\
Ryzen 7 7840U & 2.77 & 2.80 & 0.80\% \\
Intel N250    & 2.37 & 2.41 & 1.89\% \\
\bottomrule
\end{tabular}}
\end{table}

Together, these results show that \design{} turns fractional bit budgets into a consistent deployment curve across quality, memory footprint, compiler layout, throughput, and energy.

\section{Conclusion}\label{sec:conclusion}
\design{} exposes fractional-bit deployment for CPU-only LLM inference by coupling activation-aware per-channel bit allocation with compile-time layout regularization. Our compiler turns irregular mixed-precision maps into bit-homogeneous kernel blocks and propagates compatible channel orders so that layout management stays largely off the runtime path. Across three LLMs and three CPU targets, \design{} preserves fractional target budgets, improves low-bit quality by 2.4--32.1\% over prior methods at a 3b target, reduces activation reorder traffic by up to 70.8\%, and keeps measured throughput and energy within 2\% of an optimized LUT-based CPU path. Overall, \design{} turns fractional bit budgets into a consistent deployment curve across quality, memory footprint, compiler layout, throughput, and energy. The target average bit-width becomes an actionable design variable: the same $B$ that controls low-bit quality recovery also predicts peak footprint and CPU throughput. The broader lesson is that low-bit LLM efficiency on CPUs is not only a quantizer problem, but a quantizer--compiler co-design problem.

\bibliographystyle{ACM-Reference-Format}
\bibliography{References/intro, References/rw_hw_arch, References/technical_contents}

\end{document}